\documentclass[letterpaper, 10 pt, conference]{ieeeconf}  % Comment this line out if you need a4paper

\IEEEoverridecommandlockouts                              % This command is only needed if 
\usepackage{amsmath,amsfonts}
\usepackage{algorithmic}
\usepackage{algorithm}
\usepackage{array}
\usepackage[caption=false,font=normalsize,labelfont=sf,textfont=sf]{subfig}
\usepackage{textcomp}
\usepackage{stfloats}
\usepackage{url}
\usepackage{verbatim}
\usepackage{graphicx}
\usepackage{cite}
\usepackage{lipsum}
\usepackage{physics}
\usepackage{booktabs}
\usepackage{color}
\usepackage{xcolor}

\title{\LARGE \bf
Few-Shot Physics-Informed Neural Network for Shape Reconstruction of Concentric-Tube Robots
}

\author{Navid Feizi*, %~\IEEEmembership{Student member,~IEEE,},
        Filipe C. Pedrosa*, %~\IEEEmembership{Student member,~IEEE,}
        Rajni V. Patel**, %~\IEEEmembership{Life fellow,~IEEE}
         and Jagadeesan Jayender** %~\IEEEmembership{Senio member,~IEEE}
\thanks{* N. Feizi and F. C. Pedrosa contributed equally to this work.}% <-this 
\thanks{** R.V. Patel and J. Jayender are co-senior authors.}% <-this 
\thanks{This work was supported by the National Institute of Diabetes and Digestive and Kidney Diseases of the National Institutes of Health (NIH) through award number R01DK119269 and National Institute of Biomedical Imaging and Bioengineering of the NIH under award number R01EB028278 (JJ) and by the Natural Sciences and Engineering Research Council (NSERC) of Canada grant RGPIN1345 (RVP), and the Canada Research Chairs Program (RVP).}% <-this % stops a space
\thanks{Navid Feizi is with the Department of Radiology at Brigham and Women’s Hospital, and Harvard Medical School, Boston, MA 02115, USA (e-mail: nfeizi@bwh.harvard.edu).}
\thanks{Filipe C. Pedrosa is with Canadian Surgical Technologies and Advanced Robotics (CSTAR), London Health Sciences Centre (LHSC), and with the Department of Electrical and Computer Engineering, Western University, London, Ontario, Canada. (e-mail: fpedrosa@uwo.ca).}
\thanks{Rajni V. Patel is with CSTAR, and with the Department of Electrical and Computer Engineering, the Department of Surgery, the Department of Clinical Neurological Sciences, and the School of Biomedical Engineering, Western University. (e-mail: rvpatel@uwo.ca).}
\thanks{Jagadeesan Jayender is with the Department of Radiology at Brigham and Women’s Hospital, and Harvard Medical School, Boston, MA 02115, USA (e-mail: jayender@bwh.harvard.edu).}}

\begin{document}

\maketitle
\thispagestyle{empty}
\pagestyle{empty}

%%%%%%%%%%%%%%%%%%%%%%%%%%%%%%%%%%%%%%%%%%%%%%%%%%%%%%%%%%%%%%%%%%%%%%%%%%%%%%%%
\begin{abstract}
    Modeling concentric tube robots (CTRs) involves complex nonlinear continuum mechanics, and despite recent progress, physics-based models often lack an accurate representation of the experimental setups. To overcome these limitations, deep neural network-based models have been explored as alternatives with superior accuracy; however, they often overlook known mechanics, require large training datasets, and typically discard shape estimation of the robot. We present a physics-informed neural network (PINN) for kinematic modeling of a 6-DoF CTR with three pre-curved tubes that embeds the Cosserat rod differential equations and learns from few-shot observational data, balancing physics priors with data-driven fitting. PINN enables full-state estimation of shape, twist angle, torsional strain, bending moment, and orientation. Benchmark tests show a mean shape error below 1\% of the robot length and accurately recovered other kinematic states, outperforming a purely physics-based Cosserat rod model baseline while using a minimal training set. The resulting model is also computationally efficient and robust, making it well-suited for real-time control applications.
\end{abstract}

\section{INTRODUCTION}
Concentric tube robots (CTRs), are a subclass of continuum robots that consist of two or more pre-curved flexible tubes \cite{dupont2022continuum, burgner2015continuum}. Axial translation and rotation of these tubes allow active manipulation of the backbone curvatures, enabling the robot to generate complex three-dimensional shapes \cite{nwafor2023design}. Due to their slender, needle-like structure and dexterity, CTRs have been widely investigated for applications requiring navigation through tortuous pathways, particularly in the minimally invasive surgery domain, including skull base, abdominal, thoracic, and cardiac \cite{mitros2022theoretical, price2023using, burgner2013telerobotic, feizi2025design, gafford2020concentric}.

Safe navigation of a CTR in medical applications requires accurate motion planning, not only for the distal end, but also for the entire body of the robot to avoid collisions with the surrounding anatomy \cite{pedrosa2022surgical}. This demands a precise kinematic model capable of estimating the shape of the robot for a given set of actuation inputs while being computationally efficient for real-time control/planning applications.

\subsection{Related Work}
Physics-based modeling of CTRs is commonly founded on the well-established Cosserat rod theory, in which the tubes are considered inextensible and without transverse shear strain \cite{antman2005nonlinear, rucker2010geometrically, rucker2010equilibrium, dupont2010, rucker2010}. Using this, the kinematics are formulated into a boundary value problem (BVP) that reconstructs the full backbone shape of the robot. The BVP can be solved using optimization-based numerical methods \cite{keller2018numerical} to obtain the states. However, optimization-based iterative solvers are computationally demanding and highly sensitive to actuation inputs and initial conditions. This leads to inconsistent computation time, making this approach inadequate for real-time path planning applications, specifically with sampling-based techniques such as rapidly-exploring random tree, where the model is evaluated frequently \cite{niu2020path}. Furthermore, despite substantial progress, physics-based models remain limited in accuracy due to unmodeled effects, including friction, tube clearance, and nonlinear constitutive laws.

Deep neural network (DNN) are known for their capability to serve as universal function approximators. They have been used to learn the kinematics or dynamics of various systems \cite{legaard2023constructing}. If a collection of solution-labeled pairs is provided, a DNN can be trained to estimate the solution mapping through supervised learning. In \cite{grassmann2018learning}, a multilayer perceptron (MLP) was trained on \mbox{80K} data points to learn the forward kinematics of a 6-DoF CTR, predicting the distal end position and orientation from joint configurations. In \cite{grassmann2019merits} an MLP was trained on \mbox{94K} data points, showing that using quaternion/vector-pairs outperforms other representations in learning the forward kinematics. Their experimental dataset, consisting of 100K configurations, was published in \cite{grassmann2022dataset}, which we have used in this paper. In \cite{kuntz2020learning}, an MLP was trained on a \mbox{100K} dataset, consisting of the shape of the CTR detected by cameras, to reconstruct the entire backbone of a three-tube CTR, demonstrating a two-fold reduction in average error and nearly six-fold reduction in maximum error compared to the physics-based model. In \cite{liang2021learning}, an inverse kinematics was learned from robot shapes captured by cameras using \mbox{100K} data points. In \cite{jeong2025learningn}, the forward and inverse kinematics of a \mbox{4-DoF} CTR were learned employing an LSTM-MLP network structure that is able to capture the snapping behavior, where the network was trained on ~200K data points recorded in 21 hours.

DNN-based modeling approaches often yield a more accurate representation of the physical robot than physics-based methods, as they can capture uncertainties and unmodeled behaviors, which are typically neglected in physics-based formulations but are implicitly learned from experimental data \cite{kuntz2020learning, grassmann2019merits, grassmann2022dataset}. Since DNN prediction requires only a forward pass through the network \cite{feizi2025deep, bensch2024physics}, rather than an iterative optimization based solver to solve the Cosserat rod BVP, these approaches are computationally less demanding and more stable, However, standard DNN is data-hungry which often requires tens of thousands of training samples as it learns solely from observation data, neglecting the underlying physics. In general, collecting a large dataset from an experimental setup is challenging and may cause wear of the setup. More importantly, despite the backbone being essential for tasks such as path planning and follow-the-leader implementations, in most conventional DNN-based approaches, except \cite{kuntz2020learning}, the shape estimation, which demands a shape-sensing modality, increasing the difficulty of generating the dataset, is disregarded, and only the end effector of continuum robots is modeled \cite{grassmann2018learning, jeong2025learningn, feizi2025deep}.

Physics-informed neural networks (PINNs) incorporate the governing physics into the training loss function along with the observation dataset \cite{raissi2019physics}. By embedding the physics into the training process, PINNs can significantly reduce the amount of required observational data, creating a balance between data and physics-based information, or even eliminating it in fully physics-based formulations as in \cite{raissi2019physics}. PINN has been recently adapted for various robotic applications. It has been used to solve time domain ordinary differential equations (ODEs) with variable initial conditions to model the dynamics of a robotic arm \cite{nicodemus2022physics}, as well as to learn the dynamics of the end effector of a tendon-driven robot \cite{liu2024physics}. Bensch et al. \cite{bensch2024physics} demonstrated that incorporating Cosserat rod differential equations into the training loss facilitated complete shape estimation of a tendon-driven continuum robot with high accuracy in simulation, while substantially reducing computational costs compared to solving the Cosserat BVP.

% In contrast to data-hungry deep networks that typically learn only end-effector pose from large datasets, and to sensitive, iterative Cosserat BVP solvers with variable runtimes, we propose a physics-informed neural network for CTRs that embeds Cosserat-rod equations and learns from a small number of experimental observations to recover the full kinematic state (backbone shape, torsion, orientation, bending moments). On benchmarks, our model achieves <1\% mean shape error while delivering consistent per-query runtimes suitable for real-time planning, addressing a gap left by prior DL surrogates and PINN-style shape estimators developed for other continuum robot classes.

\subsection{Contributions}
% To the best of our knowledge, existing DNN models for the forward kinematics of CTRs ignore the governing physics and only use observation datasets that typically require \mbox{80-200K} samples for training. Moreover, most prior work, except \cite{kuntz2020learning}, estimates only the end-effector pose and neglects the full shape reconstruction, which is essential for path planning and follow-the-leader applications.

In this work, we propose, to the best of our knowledge, the first PINN for the real-time forward kinematics of a 6-DoF CTR in free space. The model incorporates Cosserat rod equations, effectively balancing physics-based principles with data-driven fitting, which significantly reduces the size of the required observation dataset while enabling the estimation of entire shapes, without using any shape-sensing modality. Our key contributions are:
\begin{itemize}
    \item Formulation of a PINN that incorporates the Cosserat rod equations for CTR, reducing the number of required experimental measurements from tens of thousands to a few hundred.
    \item Recovery of the entire 3D backbone shape of the CTR without the need for a shape-sensing modality.
    \item Estimation of the latent states of the CTR, including twist angle, torsional strain, and orientation.
    \item Comprehensive evaluation in simulation and with the open-source experimental dataset in \cite{grassmann2022dataset}.
\end{itemize}

\subsection{Outline}
The remainder of this paper is organized as follows. In Section~\ref{sec:preliminaries}, we introduce some preliminaries, including the Cosserat rod theory for modeling CTRs and the foundational concepts of PINNs. Section~\ref{sec:methods} describes the proposed PINNs framework and training strategy. In Section~\ref{sec:results}, we present simulation and experimental results. Finally, Section~\ref{sec:conclusions} provides concluding remarks.

\section{PRELIMINARIES}
\label{sec:preliminaries}

\subsection{Cosserat Rod Theory}
\paragraph{Kinematics}
Using Cosserat rod theory \cite{antman2005nonlinear, dupont2010, rucker2010},
the shape of a rod (tube) of length~$\ell$ is represented by a continuous homogeneous transformation \mbox{$g(s)\in SE(3)$} along the centerline of the rod, where $g(s)$ consists of a position vector \mbox{$\boldsymbol{p}(s)\in\mathbb{R}^3$} and a rotation matrix \mbox{$\mathbf{R}(s)\in SO(3)$}, where $s\in[0, \ell]$ denotes the arc-length (for notational simplicity, we omit the explicit dependence on $s$ in the following equations). Accordingly, the spatial derivatives of \mbox{$\mathbf{R}$} and \mbox{$\boldsymbol{p}$} are given by
\begin{equation}
    \dot{\mathbf{R}} = \mathbf{R}\,\widehat{\boldsymbol{u}},
    \quad
    \dot{\boldsymbol{p}} = \mathbf{R}\,\boldsymbol{v}
    \label{eq:kinematics}
\end{equation}
where \mbox{$\boldsymbol{v}$} and \mbox{$\boldsymbol{u}$} are the linear and angular strains, respectively. In other words, the vectors \mbox{$\boldsymbol{v}$} and \mbox{$\boldsymbol{u}$} describe how \mbox{$\boldsymbol{p}$} and \mbox{$\mathbf{R}$} vary with respect to the arc-length of the rod. The operator $(\widehat{\cdot})$ denotes the Lie algebra of $\mathrm{SO}(3)$, $3 \times 3$ skew-symmetric matrices mapping vectors from $\mathbb{R}^3$ to $\mathfrak{so}(3)$.

\paragraph{Constitutive Law}
Based on the linear constitutive law, the relationship of the strains to the internal force $\boldsymbol{n}$ and moment $\boldsymbol{m}$ at $s$ is described as follows:
\begin{subequations}
    \label{eq:constitutive}
    \begin{align}
        \boldsymbol{n} & = \mathbf{R}\,\mathbf{K}_{se}(\boldsymbol{v} - \boldsymbol{v}^*) \label{eq:constitutive_n} \\
        \boldsymbol{m} & = \mathbf{R}\,\mathbf{K}_{bt}(\boldsymbol{u} - \boldsymbol{u}^*) \label{eq:constitutive_m}
    \end{align}
\end{subequations}
where \mbox{$\mathbf{K}_{se} = \mathrm{diag}(GA,\, GA,\, EA)$} is the shear-extension stiffness matrix, \mbox{$\mathbf{K}_{bt} = \mathrm{diag}(EI_{xx},\, EI_{yy},\, GI_{zz})$} is the bending-torsion stiffness matrix, $A$ is the cross-sectional area, $E$ is the Young's modulus, $G$ is the shear modulus, and $I_{xx}, I_{yy}$ are the second moments of area about the principal axes of the cross-section, and $I_{zz}$ is the polar moment of inertia. The unloaded linear strain of the rod is given by \mbox{$\boldsymbol{v}^\ast = [0,~0,~1]^\top$}, and the unloaded curvature (pre-curvature) of the rod is \mbox{$\boldsymbol{u}^\ast = [\kappa_i,~0,~0]^\top$}, since we only assume planar unloaded curvature about the $x$-axis.

\paragraph{Equilibrium Equations}
The static equilibrium equations of a rod subject to a distributed external force $\boldsymbol{f}$ and moment $\boldsymbol{l}$ per unit length are obtained by integrating the free-body diagram of a cantilever beam extending from $s$ to $\ell$ \cite{antman2005nonlinear}. Differentiating the mechanical equilibrium equation with respect to the arc-length~$s$ yields the classical equilibrium equations for a Cosserat rod:
\begin{subequations}
    \label{eq:equilibrium}
    \begin{align}
        \dot{\boldsymbol{n}} + \boldsymbol{f}                                                  & = \boldsymbol{0} \label{eq:equilibrium_n} \\
        \dot{\boldsymbol{m}} + \widehat{\dot{\boldsymbol{p}}}\,\boldsymbol{n} + \boldsymbol{l} & = \boldsymbol{0} \label{eq:equilibrium_m}
    \end{align}
\end{subequations}

Differentiating \eqref{eq:constitutive_m} and using \eqref{eq:kinematics}, we can write \eqref{eq:equilibrium_m} in terms of $\boldsymbol{u}$, $\boldsymbol{f}$, and $\boldsymbol{l}$ as follows:
\begin{equation}
    \dot{\boldsymbol{u}}
    = \dot{\boldsymbol{u}}^{*}
    - \mathbf{K}^{-1} \Big( \widehat{\boldsymbol{u}}\mathbf{K}(\boldsymbol{u} - \boldsymbol{u}^{*})
    + \widehat{\boldsymbol{e}}_{3}\mathbf{R}^{\top} \int_{s}^{\ell} \boldsymbol{f}(\sigma) d\sigma
    + \mathbf{R}^{\top}\boldsymbol{l} \Big)
    \label{eq:curvature}
\end{equation}
where \mbox{$\boldsymbol{e}_3 = [0,~0,~1]^\top$}. Equation \eqref{eq:curvature} along with \eqref{eq:kinematics} forms a system of differential equations with boundary conditions at \mbox{$s=0$} and \mbox{$s=\ell$} that can be solved to reconstruct the shape of a rod under external loads.

\paragraph{Static Model of a Collection of Tubes}
Considering a CTR composed of $n$ tubes, the static equilibrium condition is obtained by summing the contributions from all tubes. Accordingly, the internal force and moment at a cross-section at arc-length~$s$ are given by the sum of the contributions from each tube as follows:
\begin{subequations}
    \label{eq:ctr_equilibrium}
    \begin{align}
        \sum_{i=1}^{n} \dot{\boldsymbol{n}}_i + \boldsymbol{f}_i                                                    & = \boldsymbol{0} \label{eq:ctr_equilibrium_n} \\
        \sum_{i=1}^{n} \dot{\boldsymbol{m}}_i + \widehat{\boldsymbol{\dot{p}}}\,\boldsymbol{n}_i + \boldsymbol{l}_i & = \boldsymbol{0} \label{eq:ctr_equilibrium_m}
    \end{align}
\end{subequations}

Assuming that the tubes are concentric, they share the same centerline curve $\boldsymbol{p}$ and tangent $\dot{\boldsymbol{p}}$, while remaining free to twist independently. Accordingly, the rotation matrix of the $i$\textsuperscript{th} tube can be expressed in terms of the rotation of the innermost tube and its relative twist angle, with respect to tube 1, $\theta_i$, as \mbox{$\mathbf{R}_i = \mathbf{R}_1 \mathbf{R}_z(\theta_i)$}, where the matrix \mbox{$\mathbf{R}_z(\theta_i)$} denotes the rotation about the $z$-axis by $\theta_i$.
\begin{equation}
    \boldsymbol{u}_i = \mathbf{R}_{z,i}^{\top}\,\boldsymbol{u}_1 + \dot{\theta}_{i}\,\boldsymbol{e}_3,
    \quad
    \dot{\theta}_{i} = u_{i,z} - u_{1,z}
    \label{eq:ui}
\end{equation}

This implies that only the $z$-components of the tube curvatures are independent variables. Using this, \eqref{eq:curvature} can be rewritten for each tube. Since the $x$ and $y$ components of the tube curvatures are identical across all tubes, we only retain the individual torsional component $u_{i,z}$ in the equations. For the sake of notational simplicity, we use the $x$ and $y$ components of the body-frame bending moment of the innermost tube ($i = 1$), $\boldsymbol{m}^{b}_{1,x,y}$, instead of $u_{1,x,y}$, leading to the following system of ODEs for an $n$-tube CTR under external loads:
\begin{equation}
    \begin{aligned}
        \dot{\boldsymbol{m}}_{1,x,y}^{b} & = \bigl(-\,\widehat{\boldsymbol{u}}_{1}\,\boldsymbol{m}^{b}_{1}\;-\;\widehat{\boldsymbol{e}}_{3}\,\mathbf{R}_{1}^{\top}\bigr)\big|_{x,y} \\
        \dot{u}_{i,z}                    & = \dot{u}^{*}_{i,z}
        + \frac{E_{i} I_{i}}{G_{i} J_{i}} \bigl(u_{i,x}\,u^{*}_{i,y} - u_{i,y}\,u^{*}_{i,x}\bigr)                                                                                   \\
        \dot{\theta}_{i}                 & = u_{i,z} - u_{1,z}                                                                                                                      \\
        \dot{\boldsymbol{p}}_{1}         & = \mathbf{R}_{1}\,\boldsymbol{e}_{3}                                                                                                     \\
        \dot{\mathbf{R}}_{1}             & = \mathbf{R}_{1}\,\widehat{\boldsymbol{u}}_{1}
    \end{aligned}
    \label{eq:ctr_ode}
\end{equation}
where
\begin{equation}
    \boldsymbol{m}^{b}_{1}
    = \begin{bmatrix} m^{b}_{1,x}, m^{b}_{1,y},  m^{b}_{1,z} \end{bmatrix}^{\top},
    \quad
    m^{b}_{1,z} = \sum_{i=1}^{N} G_{i} J_{i}\,u_{i,z}.
\end{equation}
and the torsional curvature for the innermost tube  is computed as
\begin{equation}
    \boldsymbol{u}_{1,x,y}
    = \mathbf{K}^{-1}\Bigl(\boldsymbol{m}^{b}_{1}
    + \mathbf{K}_{1}\boldsymbol{u}_{1}^{*}
    + \mathbf{R}_{2}\mathbf{K}_{2}\boldsymbol{u}_{2}^{*}
    + \mathbf{R}_{3}\mathbf{K}_{3}\boldsymbol{u}_{3}^{*}\Bigr)\Big|_{x,y}
\end{equation}

\begin{figure}[!tb]
    \centering
    \includegraphics[keepaspectratio]{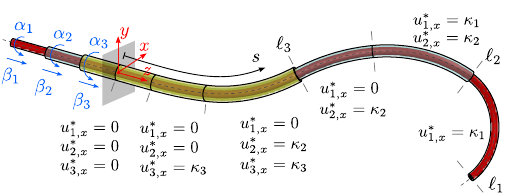}
    \caption{A three-tube CTR demonstrating six segments along its backbone as well as transition points where continuity must be maintained.}
    \label{fig:ctr_diagram}
\end{figure}

\paragraph{Boundary Conditions}
Inputs to the model are described by linear $\beta_i$ and angular $\alpha_i$ actuation values applied at the proximal end of the $i$\textsuperscript{th} tube, as shown in Fig.~\ref{fig:ctr_diagram}. Thus, the differential equations in \eqref{eq:ctr_ode} are subject to the following boundary conditions at the base ($s=0$):
\begin{equation}
    \boldsymbol{\theta}(0) = \begin{bmatrix}
        0                                                                       \\
        \alpha_{2} - \beta_{2}\,u_{2,z}(0) - (\alpha_{1} - \beta_{1}\,u_{1,z}(0)) \\
        \alpha_{3} - \beta_{3}\,u_{3,z}(0) - (\alpha_{1} - \beta_{1}\,u_{1,z}(0))\end{bmatrix}
    \label{eq:theta_bc}
\end{equation}
\begin{equation}
    \boldsymbol{p}(0) = [0, 0, 0]^\top
    \label{eq:p_bc}
\end{equation}
\begin{equation}
    % \boldsymbol{h}(0) = \text{quat}([0, \; \alpha_{1} - \beta_{1}\,u_{1,z}(0), \; 0])
    \boldsymbol{R}(0) = \text{Rot}_z(\alpha_{1} - \beta_{1}\,u_{1,z}(0))
    \label{eq:h_bc}
\end{equation}
% where $\boldsymbol{h}$ is the quaternion representation of the rotation matrix $\mathbf{R}$, which is used to avoid singularities.

The known distal boundary condition at \mbox{$s=\ell_i$ }for a CTR in free space is given by:
\begin{equation}
    \begin{aligned}
        m^b_{1,x,y}(\ell_1)     & = 0                    \\
        G_i J_i u_{i,z}(\ell_i) & = 0,  \;\; i = 1,...,n \\
    \end{aligned}
    \label{eq:bc_distal}
\end{equation}
where $\ell_i$ is the arc-length at the distal termination point of tube $i$, as shown in Fig.~\ref{fig:ctr_diagram}.

\paragraph{Implementations}
A challenge in solving the BVP defined by \eqref{eq:ctr_ode} and \eqref{eq:theta_bc} - \eqref{eq:bc_distal} arises from the discontinuities in stiffness and pre-curvature of the tubes at transition points. We divide the spatial domain into multiple segments, as show in Fig.~\ref{fig:ctr_diagram}, and ensure continuity of the position, orientation, and bending moments at the interfaces between these segments.
\begin{equation}
    g_i(s^-) = g_i(s^+), \;\;\; \;  m_i(s^-) = m_i(s^+)
\end{equation}

The BVP in~\eqref{eq:ctr_ode} is non-trivial because the constraints are split between the proximal and distal ends of the tubes. In practice, one can employ a shooting method: (i) guess the unknown proximal boundary conditions, (ii) integrate the ODE forward, and (iii) iteratively update the guess so that the distal boundary conditions are satisfied~\cite{keller2018numerical}. This can be computationally expensive and highly sensitive to initial guesses.

\subsection{Physics-Informed Neural Networks}
DNN relies solely on observational data, neglecting the underlying physics of the system, which necessitates a large training dataset to generalize effectively. In many engineering scenarios, including robotics, the underlying physics of the system is well investigated and can be included in training DNNs. By incorporating physics, the DNN is essentially restricted to a lower dimension, which allows it to be trained using a limited amount of data \cite{karniadakis2021physics}.

In the seminal work by Raissi et al. \cite{raissi2019physics} PINN was proposed for learning the solution of a partial differential equation (PDE) by incorporating the differential equation, boundary, and initial functions into the training loss. This spatio-temporal function approximation is an alternative for solving nonlinear PDEs, eliminating the need for sampled solutions for training, as well as avoiding linearization or time-stepping.

By discarding the temporal domain, PINN can also be used to solve BVPs. Assume a differential equation defined by \mbox{$\dot{\boldsymbol{x}} = \mathcal{N}(\boldsymbol{x}, s)$} with boundary function \mbox{$g(\boldsymbol{x}, s_{bound}) = 0$}, PINN approximates $\boldsymbol{x}$ for \mbox{$s \in [0, \ell]$}. The parameters of the neural network are trained by minimizing the following loss function:
\begin{equation}
    \begin{aligned}
        \mathcal{L} = & \; \lambda_{ode} \mathcal{L}_{ode} + \lambda_{bc} \mathcal{L}_{bc}                                                             \\
        =             & \;  \dfrac{1}{N_{ode}} \sum_{i=1}^{N_{ode}} \left\| \nabla_s(\boldsymbol{x_i})  - \mathcal{N}(\boldsymbol{x_i}, s_i)\right\|_2 \\
                      & +  \dfrac{1}{N_{bc}} \sum_{i=1}^{N_{bc}} \left\| g(\boldsymbol{x_i}, s_i)\right\|_2                                            \\
    \end{aligned}
    \label{eq:loss}
\end{equation}
where $\boldsymbol{x}_i$ is the PINN output at $s_i$ and $\nabla_s$ is the derivative operator with respect to $s$ that can be computed by applying the chain rule using automatic differentiation, and $N_{ode}$ and $N_{bc}$ are the number of collocation points for the ODE and boundary conditions, respectively. The terms $\lambda_{ode}$ and $\lambda_{bc}$ are weights that balance the contributions of loss terms.

\section{METHODS}
\label{sec:methods}
We consider a three-tube CTR with independently actuated translations and rotations of each tube with the design parameters mentioned in Table~\ref{tab:tube_parameters}, which is the same as the CTR used in \cite{grassmann2022dataset}. The actuation input is defined as \mbox{$\boldsymbol{\tau} = [\,\beta_{1},\,\beta_{2},\,\beta_{3},\,\alpha_{1},\,\alpha_{2},\,\alpha_{3}\,]^{\top} \in \mathcal{B}\subseteq \mathbb{R}^{6}$}, where $\mathcal{B}$ represents the allowable actuation space. The system state is represented by \mbox{$\boldsymbol{x} = [\,\boldsymbol{m}_{1,x,y}^{b},\, u_{i,z},\, \theta_{i},\, \boldsymbol{p}_{1},\, \boldsymbol{h}_{1}\,]^{\top} \subseteq \mathbb{R}^{15}$}, as described in \eqref{eq:ctr_ode}.

\begin{table}[!tb]
    \centering
    \caption{CTR Tube Kinematic Parameters}
    \label{tab:tube_parameters}
    \begin{tabular}{@{}l|ccc@{}}
        \toprule
        Parameter                       & Tube I   & Tube II  & Tube III \\
        \midrule
        Inner diameter [mm]             & $0.40$   & $0.70$   & $1.20$   \\
        Outer diameter [mm]             & $0.50$   & $0.90$   & $1.50$   \\
        Straight length [mm]            & $169.00$ & $65.00$  & $10.00$  \\
        Curved length [mm]              & $41.00$  & $100.00$ & $100.00$ \\
        Curvature ($\kappa$) [m$^{-1}$] & $28$     & $12.4$   & $4.37$   \\
        Young's Modulus ($E$) [GPa]     & $50.00$  & $50.00$  & $50.00$  \\
        Shear Modulus ($G$) [GPa]       & $19.23$  & $19.23$  & $19.23$  \\
        \bottomrule
    \end{tabular}%
\end{table}

\subsection{Implementation}
\label{sec:network_toppology}
The PINN approximates the mapping \mbox{$\boldsymbol{\varphi}: \mathbb{R}^{7} \rightarrow \mathbb{R}^{15}$}, where the inputs are arc-length and joints actuation \mbox{$[s, \boldsymbol{\tau}\,] \in [0,\ell_{1}] \times \mathbb{R}^{6}$}, and the output is the state \mbox{$\hat{\boldsymbol{x}}(s) \subseteq \mathbb{R}^{15}$}. The PINN architecture is an MLP with six hidden layers and 100 nodes per layer, and the $\tanh$ activation function was used for all layers except the output layer.

\begin{figure}[!tb]
    \centering
    \includegraphics[keepaspectratio]{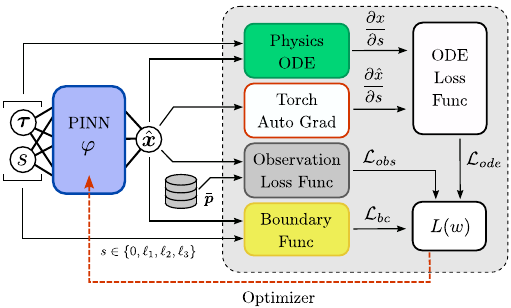}
    \caption{Block diagram of the PINN including the structure and training loss terms.}
    \label{fig:pinn_diagram}
\end{figure}

We extend the loss function \eqref{eq:loss} adding observation loss \mbox{$\mathcal{L}_{\text{obs}}$} to not only learn from the physics of the system but also from the experimental observation dataset. The physics loss enforces the PINN to learns the evolution/dynamics of the system \eqref{eq:ctr_ode}, such that $\nabla_s \hat{\boldsymbol{x}}$, the derivative of PINNs with respect to $s$, follows the system dynamics $\dot{\boldsymbol{x}}$, while the boundary and observation loss terms anchor the solution to the specified region of the state space. Fig~\ref{fig:pinn_diagram} illustrates the block diagram of the proposed PINN and the loss terms.

The ODE loss is defined as follows:
\begin{equation}
    \mathcal{L}_{\text{ode}}
    = \frac{1}{N_{\text{ode}}}
    \sum_{i=1}^{N_{\text{ode}}}
    \sum_{g \in \{m^b,\;u,\;\theta,\;p,\;h\}}
    \lambda_g^{ode} \,\big\| \boldsymbol{e}_{g,i} \big\|_2
    \label{eq:loss_phys}
\end{equation}
where
\begin{equation}
    \boldsymbol{e}^{ode}_i = [\boldsymbol{e}^{ode}_{m^b}, \boldsymbol{e}^{ode}_u, \boldsymbol{e}^{ode}_{\theta}, \boldsymbol{e}^{ode}_p, \boldsymbol{e}^{ode}_h] = \nabla_s \hat{\boldsymbol{x}}_i - \mathcal{N}(\hat{\boldsymbol{x}}_i, s_i, \boldsymbol{\tau}_i)
\end{equation}
where $\lambda_g^{ode}$ is a weighting vector for each state group, balancing the loss terms to ensure equal loss scales. $N_{\text{ode}}$ is the number of collocation points sampled in the domain \mbox{$[0,\ell_{1}] \times \mathcal{B}$}, which will be described in the next section. It should be noted that $\mathcal{N}$ is described by \eqref{eq:ctr_ode}, and the system parameters $E$, $G$, $I$, $J$, and $\boldsymbol{u}^\ast$ vary depending on the segment of the robot in which $s_i$ lies.

The boundary loss is defined as follows:
\begin{equation}
    \mathcal{L}_{\text{bc}}
    = \frac{1}{N_{\text{bc}}}
    \sum_{i=1}^{N_{\text{bc}}}
    \sum_{g \in \{m^b,\;u,\;\theta,\;p,\;h\}}
    \lambda_g^{bc} \,\big\| \boldsymbol{e}_{g,i} \big\|_2
    \label{eq:loss_bound}
\end{equation}
where $\lambda_g^{ode}$ is a weighting vector. To compute $\boldsymbol{e}_i$, four forward calls to the PINN are required at \mbox{$s_i = \{0, \ell_{3,i}, \ell_{2,i}, \ell_{1,i}\}$} with the actuation inputs $\boldsymbol{\tau}_i$, and \mbox{$\boldsymbol{e}^{bc}_i = [\boldsymbol{e}^{bc}_{m^b}, \boldsymbol{e}^{bc}_u, \boldsymbol{e}^{bc}_{\theta}, \boldsymbol{e}^{bc}_p, \boldsymbol{e}^{bc}_h]$}.
\begin{equation}
    \begin{aligned}
        \boldsymbol{e}^{bc}_{m^b}    & = \boldsymbol{x}(\ell_{1,i},\boldsymbol{\tau}_i)_{m_{x,y}}                                    \\
        \boldsymbol{e}^{bc}_{u}      & = \boldsymbol{x}(\ell_{j,i},\boldsymbol{\tau}_i)_{u_z}, \quad \text{for } j \in \{1,2,3\}     \\
        \boldsymbol{e}^{bc}_{\theta} & = \boldsymbol{x}(0,\boldsymbol{\tau}_i)_{\theta} - \boldsymbol{\theta}(0,\boldsymbol{\tau}_i) \\
        \boldsymbol{e}^{bc}_{p}      & = \boldsymbol{x}(0,\boldsymbol{\tau}_i)_{p} - \boldsymbol{p}(0,\boldsymbol{\tau}_i)           \\
        \boldsymbol{e}^{bc}_{h}      & = \boldsymbol{x}(0,\boldsymbol{\tau}_i)_{h} - \boldsymbol{h}(0,\boldsymbol{\tau}_i)           \\
    \end{aligned}
\end{equation}

In a free space case, $\boldsymbol{e}^{bc}_u$ and $\boldsymbol{e}^{bc}_m$ are zero at the boundaries, as in \eqref{eq:bc_distal} and $\boldsymbol{\theta}$, $\boldsymbol{p}$, and $\boldsymbol{h}$ are as described in \mbox{\eqref{eq:theta_bc}-\eqref{eq:h_bc}}.

The observation loss is defined as follows:
\begin{equation}
    \mathcal{L}_{\text{obs}}
    = \frac{1}{N_{\text{obs}}}
    \sum_{i=1}^{N_{\text{obs}}} \lambda_p^{obs} \,\big\| \boldsymbol{x}(s_i,\boldsymbol{\tau}_i)_{p} - \bar{\boldsymbol{p}}(s_i,\boldsymbol{\tau}_i)\big\|_2
    \label{eq:loss_obs}
\end{equation}
where $\bar{\boldsymbol{p}}$ is a few-shot position observation data, that can be obtained from  experimental measurement or the solution of the Cosserat rod model in simulation. All weighting terms $\lambda_g^{odr}$, $\lambda_g^{bc}$, and $\lambda_p^{obs}$ are tuned empirically to balance the loss terms.

\begin{figure}[!tb]
    \centering
    \includegraphics[keepaspectratio, width=0.97\linewidth]{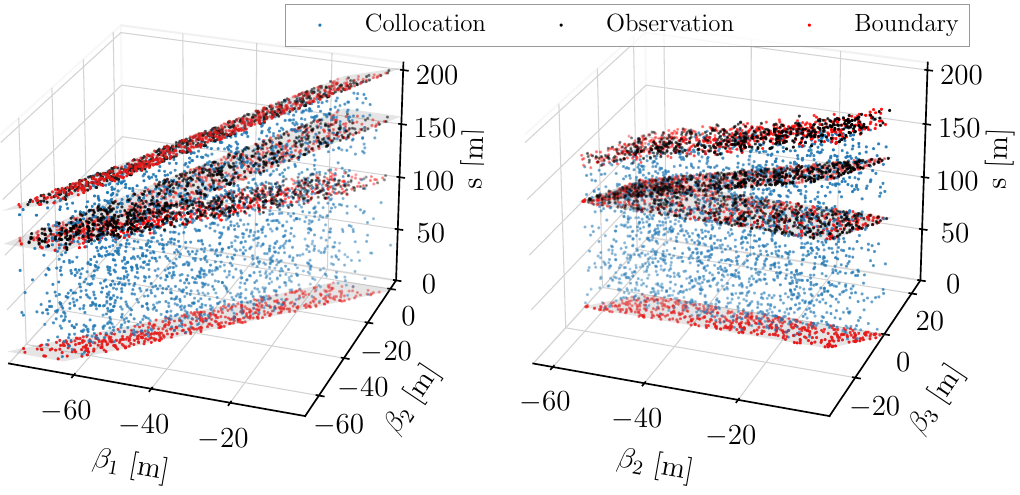}
    \caption{Linear actuation $\beta_i$ and arc-length $s$ of the training data points. Collocation, boundary, and observation samples are shown in blue, red, and black, respectively. All samples lie within the boundary defined by the inequalities for $\beta_i$. The gray regions indicate the planes at the base and distal ends of the tubes.}
    \label{fig:train_data_point}
\end{figure}

The PINN and the Cosserat rod equations were implemented in Python using PyTorch tensors, enabling using \texttt{Autograd} to compute the spatial derivative $\nabla_s \hat{\boldsymbol{x}}$ and also backpropagation through the Cosserat residuals for training.

\subsection{Training}
The collocation points for the ODE loss were sampled using a uniform random sampler in \mbox{$[0,\ell_{1}] \times \mathcal{B}$}, where $\mathcal{B}$ is primarily constrained by the physical joint limits of the robot. The boundary points were also sampled similarly for $\boldsymbol{\tau}$. However, the arc-length was set to the boundary points of the three tubes \mbox{$s = \{0, \ell_3, \ell_2, \ell_1\}$}.

\begingroup
\setlength{\parindent}{0pt}
\paragraph*{Remark} Due to the mechanical design of CTRs, specifically certain joint configurations, and particularly those involving high torsion, there may exist multiple solutions to the BVP \cite{xu2014kinematic, gilbert2015elastic}. This phenomenon, known as snapping, causes the robot to abruptly transition between equilibrium configurations. Since PINNs are not designed to handle such multi-modal behaviors \cite{raissi2019physics}, we restrict the actuation domain to avoid these scenarios. Specifically, the proximal translation of tube~3 was limited to \mbox{$-10.0 \leq \beta_{3} \leq 0.0$~mm}, the proximal translation of tube~2 to \mbox{$(\beta_{3} - 55.0) \leq \beta_{2} \leq \beta_{3}$~mm}, and the proximal translation of tube~1 to \mbox{$(\beta_{2} - 15.0) \leq \beta_{1} \leq \beta_{2}$~mm}. The rotational inputs of all tubes were constrained to \mbox{$-\pi \leq \alpha_{i} \leq \pi$}.
\endgroup

We train the PINN in two stages. (a) Simulation-based training: We generate synthetic observations with a Cosserat rod model; each observation is a set of distal-tip positions of the three tubes. Samples are drawn over the actuation space \mbox{$\boldsymbol{\tau} \in \mathcal{B}$} and at \mbox{$s = \{\ell_1, \ell_2, \ell_3\}$}. (b):~Experiment-based training: We continue training by incorporating experimental tip-position measurements for tubes 1--3 from the dataset published in \cite{grassmann2022dataset} into the observation loss $\mathcal{L}_{\mathrm{obs}}$. Note that~\cite{grassmann2022dataset} spans a broader range for $\beta_i$; however, because $\alpha_i$ are constrained to $\pm \pi/3$, the actuation space is snap free. The collocation, boundary, and observation sets for the simulation-based case are illustrated in Fig.~\ref{fig:train_data_point}, and the sample counts are listed in Table~\ref{tab:training_data_points}.

\begin{table}[!tb]
    \centering
    \caption{Training data points}
    \label{tab:training_data_points}
    \begin{tabular}{@{}c|ccc|c@{}}
        \toprule
                   & collocation & boundary & obs (model) & obs (\cite{grassmann2022dataset} ) \\
        \midrule
        \# samples & 20,000      & 1,000    & 1,000       & 500                                \\
        \bottomrule
    \end{tabular}
\end{table}

% \begin{table}[!tb]
%     \centering
%     \caption{Actuation limits}
%     \label{tab:actuation_limits}
%     \begin{tabular}{@{}c|cccc@{}}
%         \toprule
%         & $\beta_{1}[mm]$ & $\beta_{2}[mm]$ & $\beta_{3}[mm]$ & $\alpha_{i}[rad]$ \\
%         \midrule
%         min & $\beta_2 - 15.0$ & $\beta_3 - 55.0$ & $-10.0$ & $-\pi$ \\
%         max & $\beta_2$ & $\beta_3$ & $0.0$ & $\pi$ \\
%         \bottomrule
%     \end{tabular}
% \end{table}

PINNs require a larger number of epochs to train compared to conventional DNNs. This is due to three reasons: (a) Multi-objective loss terms, which may involve state terms that are scaled very differently, sometimes conflicting. $\mathcal{L}_{bc}$ and  $\mathcal{L}_{obs}$ often converge quickly at first, then the optimizer needs tens of thousands of steps to minimize the $\mathcal{L}_{ode}$. While using adjusted loss weights, as described in Section~\ref{sec:network_toppology}, provides some balance, it is insufficient, leading to gradient pathologies \cite{wang2021understanding}. (b) Ill-conditioned ODE loss: Higher-order derivatives from autograd make the loss landscape poorly conditioned, so the optimizer takes many steps to balance terms \cite{wang2021understanding}. (c) The spectral bias of the network: PINNs tend to capture the low-frequency components of the solution first, while higher-frequency components require substantially more training epochs to be learned \cite{rahaman2019spectral}. All these together make training to require multiple steps, and highly sensitive to the optimizer.

For training, we used the L-BFGS \cite{liu1989limited} optimizer, which approximates second-order curvature for optimization, making it robust to the loss landscapes conditions. In our experiments, L-BFGS significantly outperformed Adam \cite{kingma2014adam}, which aligns with recent findings in \cite{urban2025unveiling} where it was observed that L-BFGS can reduce the PINN loss by several orders of magnitude more than Adam, attributed to the ill-conditioning of the loss landscape, which a first-order gradient-based method cannot address. We initialized the network weights using Xavier \cite{glorot2010understanding} and trained the network with learning rate $2.0$, tolerance  1\texttimes{}10\textsuperscript{-10}, a ``strong-Wolfe" line search method. The training was performed on a workstation with an NVIDIA RTX 3080 GPU for enough epochs $ \geq 300,000$ until the error reduced to an acceptable level.

\section{RESULTS AND DISCUSSION}
\label{sec:results}

\subsection{Trained using Synthesized Observations Dataset}
\label{sec:res_synthesized}
\begin{figure}[!tb]
    \centering
    \includegraphics[keepaspectratio]{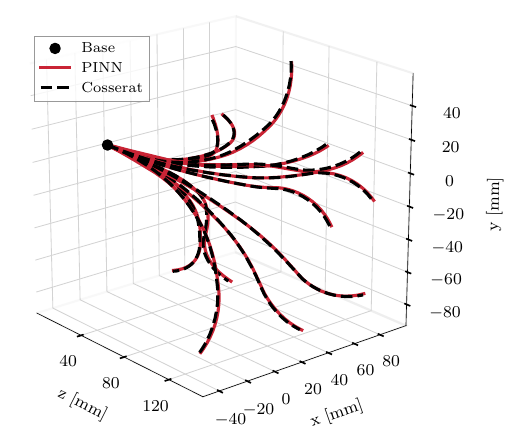}
    \caption{Shape estimation of the PINN trained on simulated observation dataset and Cosserat rod model for random actuations.}
    \label{fig:shape_sim}
\end{figure}

\begin{figure}[!tb]
    \centering
    \includegraphics[keepaspectratio]{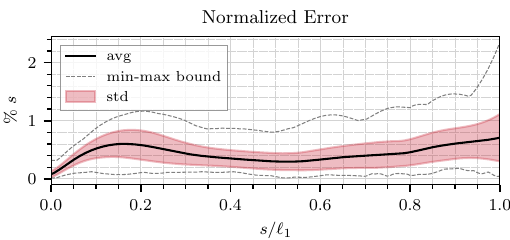}
    \caption{Average backbone error with standard deviation and minimum-maximum bounds. The Euclidean distance error was normalized by the corresponding arc-length along the backbone. The horizontal axis is normalized by $\ell_1$.}
    \label{fig:error_norm}
\end{figure}

Fig.~\ref{fig:shape_sim} shows the three-dimensional shape of the CTR as estimated by the PINN, in comparison to the Cosserat rod model, which serves as the ground truth, using 12 randomly selected unseen actuation inputs. Fig.~\ref{fig:error_norm} shows the distribution of the error normalized with respect to the arc-length for 100 randomly selected unseen actuation inputs. The Euclidean distance error for each point along the backbone was normalized by the arc-length $s$. Since the length of the CTR, $\ell_1$, varies based on $\beta_1$, the arc-length axis was normalized by $\ell_1$. The results indicate that the PINN estimated the shape with an error margin of less than 1\% along the body of the robot, and a maximum error of less than 2.5\%, closely matching the Cosserat rod model.

\begin{figure}[!tb]
    \centering
    \includegraphics[keepaspectratio]{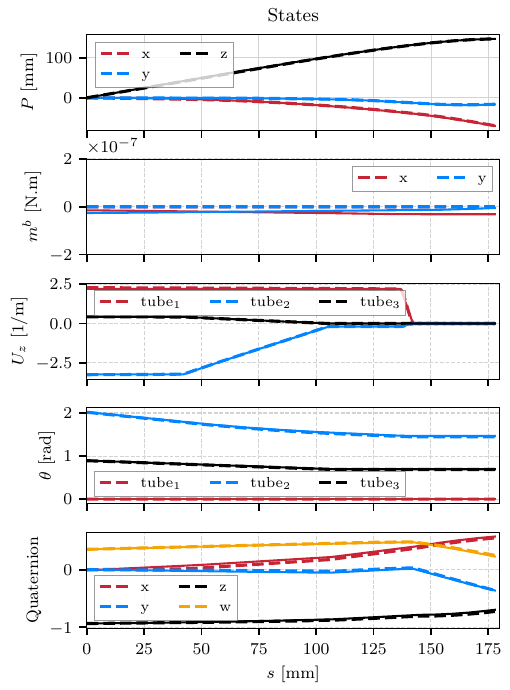}
    \caption{States estimations of the PINN (dashed lines) and the Cosserat rod model (solid lines - not mentioned in the legends) vs. arc-length for a random actuation input.}
    \label{fig:res_states}
\end{figure}

Fig.~\ref{fig:res_states} shows the states predicted by the trained PINN and the solution of the Cosserat rod model for an unseen actuation input \mbox{$\tau = [-0.032, -0.023, -0.005, -2.49, -0.32, -1.52]$}. The results demonstrate that the PINN accurately predicts the states of the CTR along its backbone. Since the modeling assumes free space, $m^b_{x,y}$, representing the bending moment of the composite tubes, is expected to be zero, which is accurately estimated by the PINN. Therefore, for the sake of brevity in the free-motion case, $m^b_{x,y}$ could be removed from the state vector. Additionally, $\theta_1$, representing the twist of tube 1 with respect to itself, is always zero and could also be removed from the state vector. However, in this work, we retain the full state representation for completeness.

\subsection{Trained using Experimental Observations Dataset}
\begin{figure}[!tb]
    \centering
    \includegraphics[keepaspectratio]{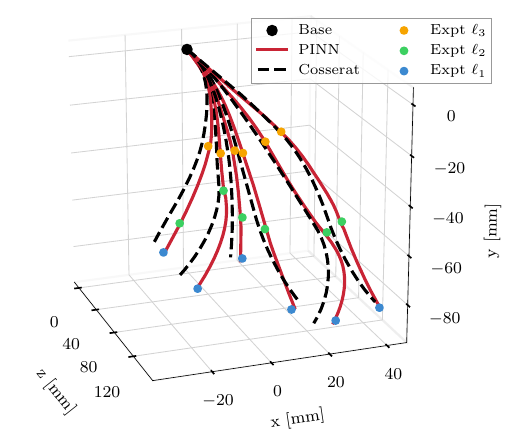}
    \caption{Shape estimation of the PINN, the Cosserat rod model, and experimental measurements of the terminal points of the tubes. The PINN was trained using 500 experimental observation data points.}
    \label{fig:shape_expt}
\end{figure}

\begin{figure}[!tb]
    \centering
    \includegraphics[keepaspectratio]{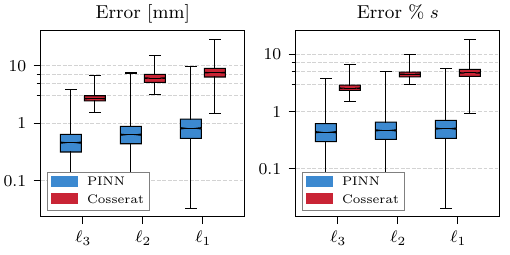}
    \caption{Error distributions at the distal terminal points of the tubes. Left: Euclidean distance error. Right: Euclidean error normalized by tube arc-length. Both panels compare PINN and Cosserat estimations.}
    \label{fig:error_expt}
\end{figure}

Fig.~\ref{fig:shape_expt} shows the backbone shapes for six randomly selected actuation inputs estimated by the pretrained PINN as described in \ref{sec:res_synthesized} and additionally trained using 500 random experimental observation data points from \cite{grassmann2022dataset}. The colored scatter points represent the experimental position measurements of the terminal points of the three tubes. Fig.~\ref{fig:error_expt} shows the distribution of position errors and normalized position errors at the terminal points for the PINN and Cosserat models across 2000 unseen randomly selected samples from \cite{grassmann2022dataset}. As can be seen in Fig.~\ref{fig:shape_expt}, the shape reconstruction using the Cosserat rod exhibits slight deviations from the experimental measurements due to mismatches in the calibrated Cosserat rod model and the physical CTR setup. In contrast, estimations of the PINN at the terminal points align closely with the experimental data while estimating the overall shape. This results in an average position estimation error of less than 1\% of the arc-length.

It is important to note that among the 2,000 randomly selected datapoints from \cite{grassmann2022dataset}, five samples exhibiting remarkably high errors at the tip positions, both for the PINN and Cosserat models, were excluded from the error plot. It is believed that these samples were outliers in the experimental dataset, likely due to measurement errors or unexpected robot behavior, such as snapping.

\subsection{Runtime Time}
\begin{figure}[!tb]
    \centering
    \includegraphics[keepaspectratio]{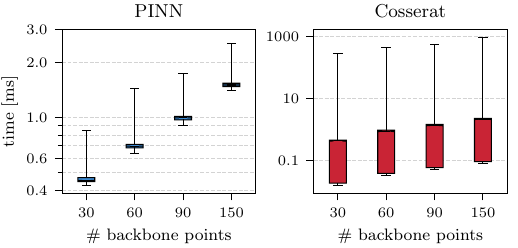}
    \caption{Runtime of the PINN and Cosserat rod model for \mbox{5,000} random actuations.}
    \label{fig:compute_time}
\end{figure}

Fig.~\ref{fig:compute_time} shows the runtimes of the PINN and the Cosserat rod model over 5{,}000 random actuation inputs. The PINN was implemented in C++ using \texttt{LibTorch}, and the Cosserat model was implemented in C++ with a Shooting-based BVP solver that uses an 8\textsuperscript{th} order Adams-Bashforth-Moulton integrator from the \texttt{Boost} library coupled with a modified Newton-Raphson method. All code was built in Release mode with GCC~13.3.0 using \texttt{-O3} optimizations, and experiments were run on a workstation with an Intel\textsuperscript{\textregistered} Core\texttrademark{} i9-10940X CPU and 256~GB RAM with isolated cores and CPU affinity (via \texttt{isolcpus} and \texttt{taskset}).

Simulations were performed for four backbone discretizations. While the median runtimes of the PINN and the Cosserat rod model do not show a remarkable difference, their variation significantly differs. PINN shows negligible variation across all configurations, whereas the Cosserat rod model shows a substantial increase in spread. The PINN runtime is almost insensitive to the actuation and arc-length, since each evaluation is a forward pass of the MLP. However, the Cosserat solver employs a shooting method that is sensitive to both actuation and initial guess at the proximal boundary. The number of iterations required to converge can vary widely, resulting in a broad range of computation times, sometimes exceeding the upper bound observed for the PINN (e.g., high torsion or large actuations) and sometimes falling below its lower bound (e.g., low torsion or small actuations). Consequently, PINN provides more consistent runtime, which is essential for real-time applications.

\subsection{Limitations}
The proposed approach demonstrates superior performance in terms of both accuracy and runtime when compared to the Cosserat rod model. However, it is limited to configurations that yield unique solutions, making it unsuitable for capturing snapping behavior. Future work will aim to address this limitation by implementing a multi-agent PINN architecture. Also, although a very limited dataset is sufficient for training, this training process is time-consuming and must be conducted for each specific CTR parameter design. Future work will address this limitation by developing a foundational model that can adapt to a variety of CTR designs rather than just one.

\section{CONCLUSIONS}
\label{sec:conclusions}
In this work, we propose a PINN to model the kinematic of a three-tube CTR in free space. The PINN integrates the Cosserat rod differential equations and boundary conditions, together with a very small dataset of only 500 experimental observations of the distal tip of the CTR, to learn the mapping from actuation to the full latent state, including the backbone shape. Simulation results showed an average shape error of less than 1\% along the backbone, and the experimental results using the dataset in \cite{grassmann2022dataset} showed that the mean position error is below 1\% at the terminal points of the three tubes. Furthermore, the proposed approach could enhance runtime efficiency compared to the traditional Cosserat rod model, making it more suitable for real-time control and planning.

\bibliographystyle{ieeetr}
\bibliography{bibliography}

\end{document}